\documentclass{article} 
\usepackage{iclr2027_conference,times}

\usepackage{amsmath,amsfonts,bm}

\def\eqref#1{equation~\ref{#1}}

\def\1{\bm{1}}

\DeclareMathAlphabet{\mathsfit}{\encodingdefault}{\sfdefault}{m}{sl}
\SetMathAlphabet{\mathsfit}{bold}{\encodingdefault}{\sfdefault}{bx}{n}

\usepackage{hyperref}
\usepackage{url}
\usepackage{algorithmic}
\usepackage{algorithm}
\usepackage{graphicx}
\usepackage{booktabs}
\usepackage[table]{xcolor}
\usepackage{amssymb}
\usepackage{makecell}
\usepackage{multirow}
\usepackage{subcaption}

\title{AdvFidelity: Unifying Preference Alignment and Few-Step Image Generation within Reinforcement Learning}

\author{
\hspace{1em}
Sicong Kang$^{1,*}$  \hspace{0.8em}
Fangao Zeng$^{1,*,\dagger}$  \hspace{0.8em}
Yuhan Li$^{2,*}$  \hspace{0.8em}
Hao Zhou$^{1}$  \hspace{0.8em}
Mengfei Xu$^{1}$   \vspace{-2mm} \AND\vspace{2mm} 
\hspace{4em}
Wenxiang Shang$^{1}$ \hspace{1em}
Wei Li$^{1}$ \hspace{1em}
Bingbing Ni$^{2}$ \hspace{1em}
Pipei Huang$^{1}$  \\
\vspace{1mm}
\hspace{32mm}
$^*$Equal Contribution \quad
$^\dagger$Project Leader \quad
\\
\hspace{13mm}
$^1$Taobao \& Tmall group of Alibaba \vspace{1mm}
$^2$Shanghai Jiao Tong University \quad
\vspace{-6mm}
}

\iclrfinalcopy 
\begin{document}

\maketitle


\begin{abstract}
Practical image generation requires both human preference alignment and few-step efficiency, yet combining reinforcement learning (RL) with distillation is often multi-stage and may erode RL gains We observe that few-step RL progressively improves structural coherence, even with specialized rewards, but leaves low-level artifacts that these rewards may fail to penalize. To address this gap, we propose AdvFidelity, which integrates artifact-sensitive fidelity supervision into RL through an online adversarial reward. This enables joint reward alignment and high-fidelity few-step generation within a single RL stage, without distribution matching distillation (DMD) or other explicit distillation objectives. We train a discriminator to rank high-quality references (positives) above content-matched reconstructions (negatives) produced by the current generator. Reconstructing moderately noised references largely preserves style and scene layout while exposing low-level degradations. This encourages the discriminator to assess artifacts rather than content differences. With negatives reconstructed online, alternating updates adapt fidelity supervision to the generator's evolving failure modes. References can come from existing target-domain data or full-step model samples. Built on LeapAlign, AdvFidelity backpropagates gradients from both rewards through the few-step sampling trajectories used at inference. Across two model backbones and three task settings, AdvFidelity matches full-step LeapAlign in alignment, task performance, and overall generation quality and remains highly competitive with other evaluated full-step RL methods, using only 4 or 8 steps without CFG.
\end{abstract}
   

\section{Introduction}

Few-step distillation and reinforcement learning (RL) are two important procedures in the post-training of image generation models. In real-world production, they are usually performed sequentially: RL is first applied for human preference alignment~\citep{flowgrpo,awm,diffusionnft,dancegrpo}, and few-step distillation is subsequently performed for efficient inference~\citep{cm,ctm,lcm,dmd2}. This sequential pipeline is cumbersome and often suboptimal, as distillation may wash out the reward gains obtained from RL. Recent works, such as DMDR~\citep{dmdr} and RTDMD~\citep{rtdmd}, adopt a joint training approach: they first warm up with distribution matching distillation (DMD)~\citep{dmd} and then optimize the RL and DMD objectives in parallel.

During joint optimization, these methods use the pretrained base model, which is not aligned with the target reward, as a fixed reference score model for DMD. Matching this base distribution may oppose the policy's shift toward reward-preferred outputs. This raises a practical question: does introducing DMD compromise the gains achieved by the same RL method in the full-step setting? Existing evaluations primarily compare joint methods with base models and other few-step generators, with limited comparisons against corresponding full-step RL baselines.

\begin{figure}[ht]
\begin{center}
\begin{subfigure}[t]{\linewidth}
\centering
\includegraphics[width=\linewidth]{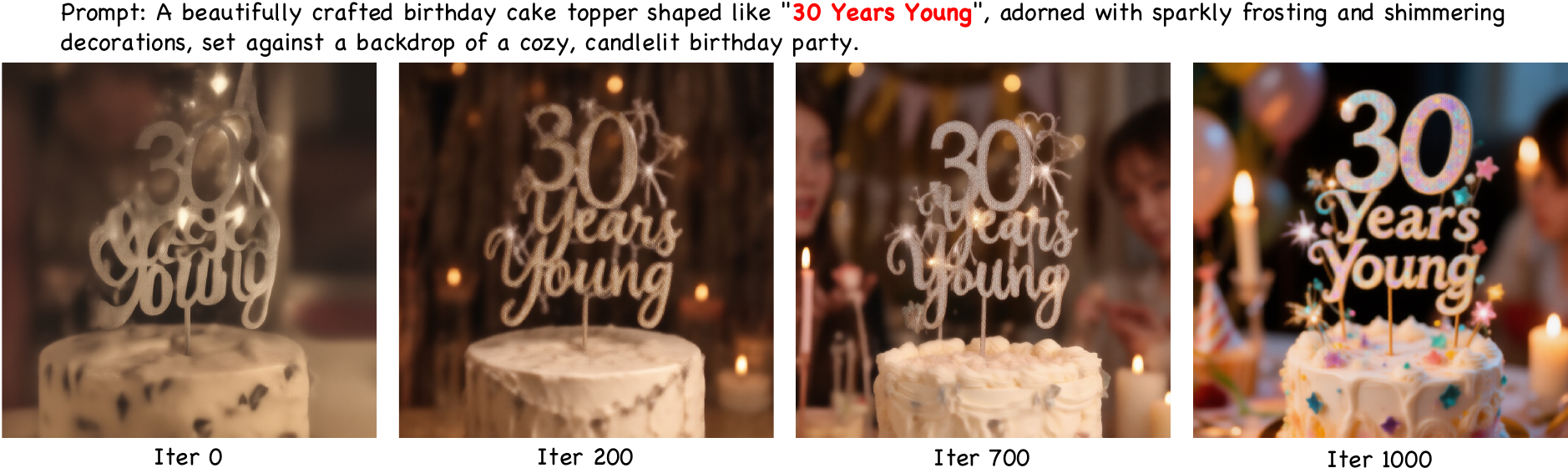}
\caption{Few-step RL with a sparse OCR reward: collapsed generations regain structural coherence during training, while grid-pattern artifacts persist after the reward is satisfied}
\label{fig:motivation-a}
\end{subfigure}
\begin{subfigure}[t]{\linewidth}
\centering
\includegraphics[width=\linewidth]{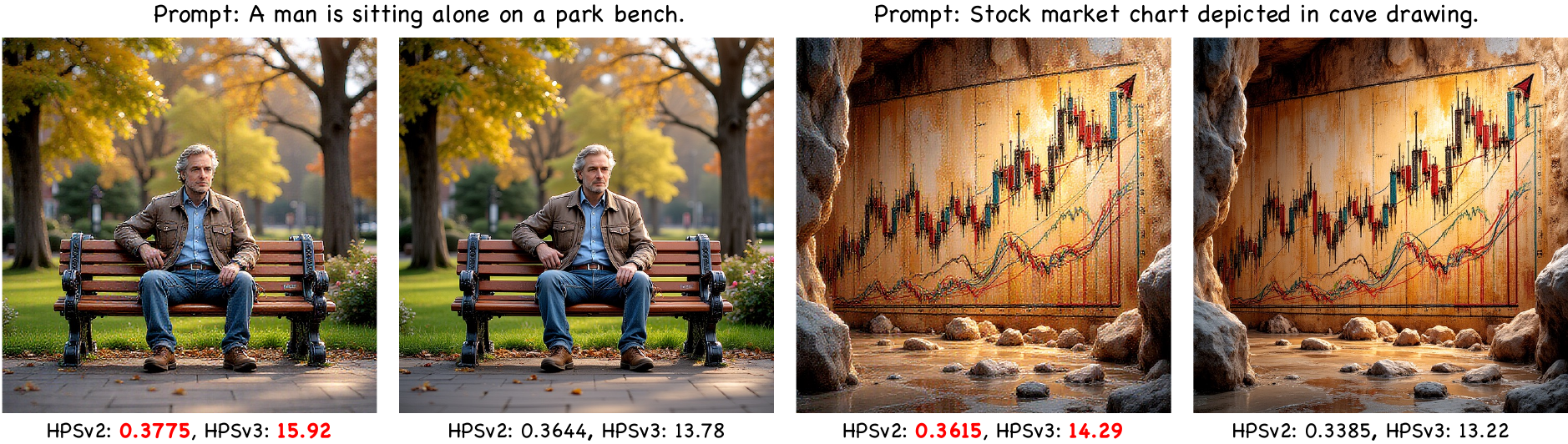}
\caption{Artifact-laden generations (left of each pair) receive preference scores comparable to or even higher than those of clean images, providing insufficient supervision to recover the clean image distribution.}
\label{fig:motivation-b}
\end{subfigure}
\end{center}
\caption{Motivation of AdvFidelity.}
\label{fig:motivation}
\end{figure}

In this work, we remove the DMD constraint of matching a frozen, RL-agnostic reference distribution. Our framework jointly learns reward alignment and few-step generation within a single RL stage, matching the alignment performance and overall generation quality of its full-step RL baseline without a separate distillation warm-up. We build on a key observation: RL pipelines that directly optimize few-step rollouts~\citep{diffusionnft,leapalign} already improve few-step image quality during training, even with highly specialized rewards (e.g., the OCR reward in Figure~\ref{fig:motivation-a}). However, these gains are incomplete: few-step outputs retain subtle noise artifacts that task rewards may fail to penalize (Figure~\ref{fig:motivation-b}), especially at later training iterations. This motivates an artifact-sensitive fidelity reward that complements task-specific rewards, enabling joint reward alignment and artifact suppression.

We propose AdvFidelity, which uses an adversarially trained discriminator~\citep{gan} as a fidelity reward. Its design addresses two practical challenges. First, a fixed discriminator may miss artifacts that emerge as the policy evolves. We therefore update the generator and discriminator alternately, training the discriminator on high-quality references and negatives reconstructed by the current generator to adapt fidelity supervision to changing failure modes. Second, differences in style, semantics, or layout can provide shortcuts for discrimination. We construct content-matched pairs by moderately noising each reference and reconstructing it with the current generator, largely preserving its high-level appearance. This reduces high-level differences and encourages the discriminator to focus on low-level artifacts.

Fidelity references can come from full-step base-model outputs, existing target-domain images, or samples from a model that covers the desired visual domain. Base-model outputs are often suitable for rewards targeting subject identity preservation, text accuracy, or compositional correctness, which improve task correctness without substantially changing the desired visual appearance. High-quality real images can also serve directly as references in industrial applications such as e-commerce and film production, where photorealism is a priority.

We instantiate AdvFidelity with LeapAlign~\citep{leapalign} and evaluate FLUX.1-dev with the HPSv2 reward on HPDv2 and GenEval prompts, and Qwen-Image-Edit-2511 on virtual dressing with a DINO similarity reward that encourages garment consistency and realism. Using only 4 or 8 sampling steps without classifier-free guidance (CFG), AdvFidelity matches full-step LeapAlign in alignment, task performance, and overall generation quality, and remains highly competitive with other evaluated full-step RL methods. The LeapAlign comparison uses the same base model, reward, and task data, with gains on multiple metrics. These settings use target-domain model sampling, base-model sampling, and target-data reuse, respectively.

Our contributions are summarized as follows.
\begin{itemize}
    \item We propose AdvFidelity, a framework that jointly learns reward alignment and few-step generation within a single RL stage, without distillation warm-up or parallel DMD training. It removes DMD's constraint of matching a frozen, RL-agnostic base distribution.
    \item We develop an online adversarial fidelity reward trained on content-matched positive--negative pairs. Online updates adapt the discriminator to evolving artifacts, while matched reconstruction reduces high-level differences and encourages the discriminator to focus on low-level quality.
    \item Experiments across two model backbones and three task settings demonstrate that, using only 4 or 8 sampling steps without CFG, AdvFidelity matches full-step LeapAlign in alignment, task performance, and overall generation quality, and remains highly competitive with other evaluated full-step RL methods.
\end{itemize}
 

\section{Related Work}
\label{related}

\noindent \textbf{Reinforcement Learning.}
Reinforcement learning has become a major approach for aligning image generation models with human preferences and task-specific criteria. Policy-gradient methods formulate denoising as a Markov Decision Process and optimize terminal rewards. DDPO~\citep{ddpo} establishes this paradigm, and Flow-GRPO~\citep{flowgrpo}, DanceGRPO~\citep{dancegrpo}, and MixGRPO~\citep{mixgrpo} adapt group-relative policy optimization~\citep{grpo} to diffusion and flow models. DPO-based methods~\citep{dpo,diffusiondpo,spo} instead learn from preference pairs without an explicit reward model. AWM~\citep{awm} and DiffusionNFT~\citep{diffusionnft} use reward-dependent objectives on forward-noised samples. Direct-gradient methods, including ReFL~\citep{imagereward}, DRaFT~\citep{draft}, and DRTune~\citep{drtune}, propagate differentiable reward gradients through sampling, with different gradient truncation and trajectory designs. LeapAlign~\citep{leapalign} uses leap trajectories to extend reward gradients to arbitrary generation steps. In pipelines such as DiffusionNFT and LeapAlign, reduced-step rollouts or shortened gradient paths primarily serve to reduce training cost, rather than explicitly train a few-step inference model. We study how fidelity supervision can support reward alignment and few-step generation at the inference sampling budget.

\noindent \textbf{Few-Step Distillation.}
Few-step distillation reduces the sampling budget of image generation models to one or a few steps. Consistency-based approaches, including Consistency Models~\citep{cm}, Latent Consistency Models~\citep{lcm}, and Phased Consistency Models~\citep{pcm}, promote consistent predictions across noise levels or trajectory phases. Distribution Matching Distillation (DMD)~\citep{dmd} estimates distribution-matching gradients using a fixed reference score model and an online fake-score model. DMD2~\citep{dmd2} removes the original regression loss, improves fake-score tracking, and incorporates adversarial supervision from real images. These methods learn few-step generation through explicit consistency or distribution-matching objectives. Applying distillation after RL may reduce reward gains. AdvFidelity instead learns few-step generation through task and fidelity rewards within RL, without an explicit distillation objective.

\noindent \textbf{Joint Reward Alignment and Few-Step Generation.}
DMDR~\citep{dmdr} and RTDMD~\citep{rtdmd} combine reward optimization with DMD after a distillation cold start, retaining fixed-reference score supervision and an online fake-score model. Their evaluations mainly compare with base models and other few-step generators. DMDR includes full-step ReFL comparisons, but its GRPO variant lacks a corresponding baseline. RTDMD reports near-parity with a published full-step Flow-GRPO baseline on GenEval, but does not establish that their training data and reward configurations are comparable, or report results on other prompt distributions for that model. Although it also uses HPS and PickScore as training rewards, it does not report corresponding full-step RL comparisons in these preference-alignment settings. These comparisons leave preservation of full-step RL gains insufficiently established.

Reward-Instruct~\citep{rewardinstruct} learns few-step generators through regularized reward maximization without diffusion distillation losses. It reports quality metrics and full-step ReFL and DRaFT comparisons, but not its own full-step counterpart. It maintains quality through parameter regularization, stochastic sampling, and fixed pretrained rewards. AdvFidelity learns an online fidelity reward and compares against full-step LeapAlign, its underlying RL method.

\noindent \textbf{Online Adversarial Rewards.}
Adv-GRPO~\citep{advgrpo} trains an adversarial reward to mitigate reward hacking and align generations with reference domains; we follow its use of a frozen DINOv2~\citep{dinov2} backbone and a trainable head. Its reported setup contrasts Qwen-Image~\citep{qwen-image} references with SD3 rollouts without constructing content-matched reference--reconstruction pairs. Consequently, fidelity differences are confounded with source-model characteristics, making style or content differences available as discrimination shortcuts. AdvFidelity combines task and fidelity rewards without DMD and constructs each negative by reconstructing a moderately noised reference with the current generator (Sec.~\ref{sec:pairs}). This largely preserves style and scene layout, reducing high-level shortcuts and encouraging the discriminator to focus on low-level artifacts. Online updates adapt this supervision as the policy evolves. Our ablations support the benefits of both adaptive fidelity supervision and content-matched negatives (Sec.~\ref{sec:exp-ablation}).
  

\section{Method}

\subsection{Preliminaries: RL on Few-Step Rollouts}
\label{sec:prelim}

We consider text-conditioned flow-matching generators that transport a noise sample $\epsilon \sim \mathcal{N}(0, I)$ to an image conditioned on a prompt $c$. With a sampling budget of $\tau$ denoising steps, the resulting policy is written as $x \sim \pi_\theta(\cdot \mid c; \tau)$. We use text-conditioned notation for simplicity; in image editing tasks, the conditioning also includes the input images.

RL-based post-training optimizes the policy against a reward $r(x, c)$ encoding human preferences or task-specific criteria:
\begin{equation}
\max_{\theta} \; \mathbb{E}_{c \sim \mathcal{C}} \, \mathbb{E}_{x \sim \pi_\theta(\cdot \mid c; \tau)} \big[ r(x, c) \big].
\label{eq:rl}
\end{equation}
Advantage-based methods, such as GRPO~\citep{grpo}, sample groups of rollouts and normalize their rewards into advantages. Reward-feedback (ReFL) methods~\citep{refl,leapalign} instead use differentiable rewards to backpropagate gradients through the sampling process. We instantiate AdvFidelity with LeapAlign~\citep{leapalign}, a ReFL-family framework that propagates reward gradients through leap-based trajectories. The generator is trained using the few-step sampling budget used at inference.

\begin{figure}[t]
\begin{center}
\includegraphics[width=1\linewidth]{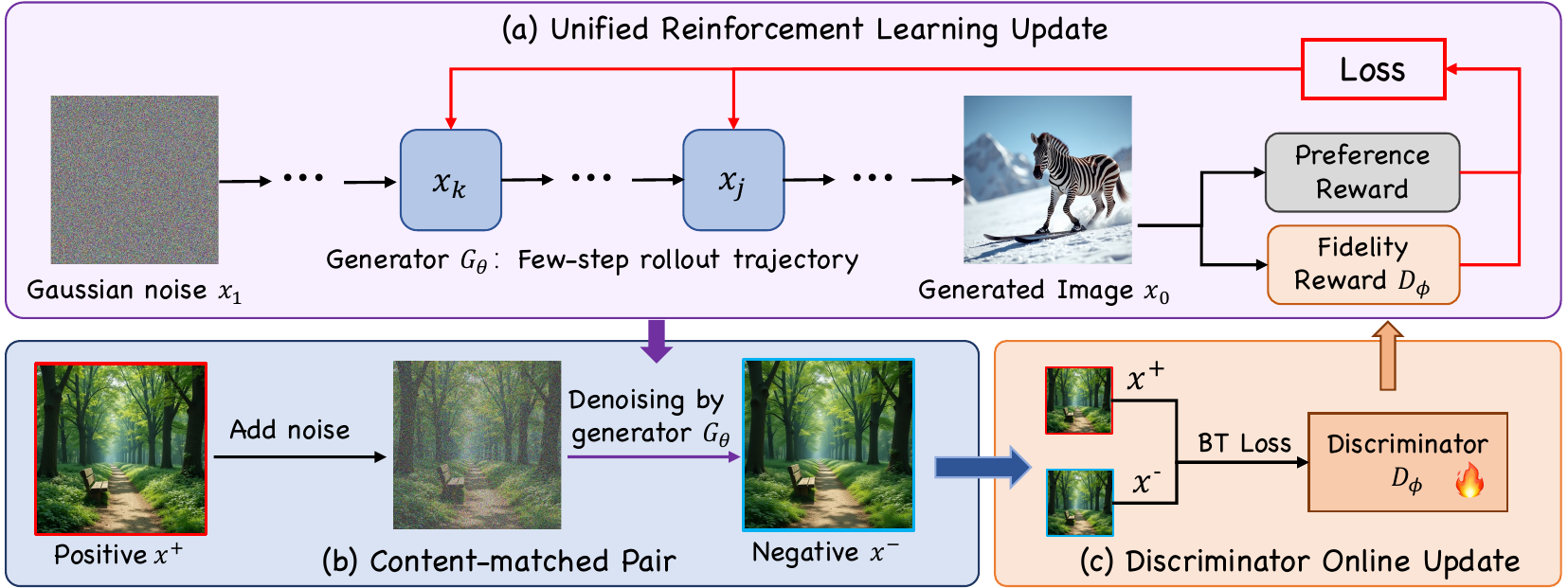}
\end{center}
\caption{\textbf{Framework of AdvFidelity.} (a) Joint RL update: the few-step generator $G_\theta$ ($\pi_\theta$) rolls out a trajectory from Gaussian noise $x_1$ to the generated image $x_0$, and is updated with both the task reward $r_{\mathrm{spec}}$ and the fidelity reward $D_\phi$. (b) Content-matched pair: a high-quality reference image $x^{+}$ is perturbed with moderate noise and reconstructed by the current $G_\theta$ to form a matched negative $x^{-}$. (c) Online update: the discriminator $D_\phi$ is trained on these pairs with a BT loss and supplies the fidelity reward for subsequent generator updates. Each training iteration performs (a) once and then repeats pair construction and the discriminator update in (b)--(c).}
\label{fig:framework}
\end{figure}

\subsection{Overview of AdvFidelity}
\label{sec:overview}

AdvFidelity builds on the observation that reward optimization on few-step rollouts can progressively recover structural coherence, while low-level artifacts persist under preference rewards alone. To supply the missing fidelity supervision, we augment the preference or task-specific reward with a fidelity reward $r_{\mathrm{fid}}$ learned by an online discriminator:
\begin{equation}
\max_{\theta} \; \mathbb{E}_{c \sim \mathcal{C}} \, \mathbb{E}_{x \sim \pi_\theta(\cdot \mid c; \tau)} \big[ r_{\mathrm{spec}}(x, c) + \lambda \, r_{\mathrm{fid}}(x) \big],
\label{eq:total}
\end{equation}
where $r_{\mathrm{spec}}$ measures human preference or task-specific criteria, $r_{\mathrm{fid}}(x)=D_\phi(x)$ assesses visual fidelity, and $\lambda$ balances the two rewards. The fidelity reward is non-stationary because the discriminator is updated alongside the policy (Sec.~\ref{sec:reward}). Under ReFL, we follow LeapAlign~\citep{leapalign} and minimize the thresholded reward loss $\mathcal{L}(\theta)=\mathbb{E}\big[\max(0,\,m-r(x,c))\big]$, with $r(x,c)=r_{\mathrm{spec}}(x,c)+\lambda D_\phi(x)$. This loss suppresses further reward updates for samples whose combined reward reaches the margin $m$.

Figure~\ref{fig:framework} illustrates the alternating generator and discriminator updates. The discriminator learns from high-quality reference images and content-matched negatives reconstructed by the current generator. References may come from existing data or be generated as needed; their selection depends on the visual requirements of the target task (Sec.~\ref{sec:pairs}). Starting from a pretrained generator, AdvFidelity jointly learns preference alignment and high-fidelity few-step generation within a single RL stage, without a separate distillation warm-up or a teacher-provided score or trajectory objective.

\subsection{Adversarial Fidelity Reward}
\label{sec:reward}

\textbf{Discriminator as reward.} Following \citet{advgrpo}, $D_\phi$ consists of a frozen DINOv2~\citep{dinov2} backbone and a lightweight trainable MLP head $h_\phi$. The head operates on the [CLS] token, without directly using patch tokens, and input images are evaluated at the RL resolution. The frozen backbone provides a stable feature space, while the online-trained head adapts to the generator's artifacts. Its score is used directly as the fidelity reward, $r_{\mathrm{fid}}(x)=D_\phi(x)$.

\textbf{Training the discriminator online.} We train $D_\phi$ to rank a high-quality reference $x^{+}$ above its matched reconstruction $x^{-}$, produced by the current policy (Sec.~\ref{sec:pairs}), using a Bradley--Terry (BT) loss:
\begin{equation}
\mathcal{L}_D = \mathbb{E}_{(x^{+}, x^{-})} \big[ {-}\log \sigma \big( D_\phi(x^{+}) - D_\phi(x^{-}) \big) \big],
\label{eq:bt}
\end{equation}
where $\sigma$ is the sigmoid function. Because the [CLS] token is a global representation, whole-image scores may be insensitive to localized defects such as grid patterns and texture noise. We therefore apply the same [CLS]-based scoring head to whole images and randomly cropped regions during generator and discriminator training. Cropping makes local defects occupy more of the input and can reduce reliance on global cues such as overall tone or border effects, without introducing a separate patch-token scorer.

\textbf{Adapting to evolving artifacts.} Generator updates can change the artifacts encountered by the discriminator. We therefore regenerate negatives with the current policy and perform multiple discriminator updates per generator update. The update ratio is specified in the implementation details. During generator updates, the discriminator parameters are held fixed while gradients propagate through its score; during discriminator updates, only the MLP head is optimized. This alternating process allows the fidelity reward to adapt as the generator changes, reducing the risk that new artifacts escape the supervision of a fixed reward.

\textbf{Compatibility with other RL methods.} The fidelity reward is a scalar and can be incorporated into different RL objectives. In our ReFL implementation, its gradient is propagated through the sampling process together with that of $r_{\mathrm{spec}}$. In advantage-based methods such as GRPO, the same score can contribute to the rewards used to compute rollout advantages. Our experiments use the LeapAlign implementation.

\subsection{Content-Matched Pair Construction}
\label{sec:pairs}

The BT loss learns any features that consistently distinguish positives from negatives. If unrelated image populations are used, differences in style, semantics, or layout can become shortcuts for discrimination. We instead construct content-matched pairs by reconstructing each reference with the current generator, largely preserving its high-level appearance while exposing low-level degradations.

\textbf{Noise-anchored negatives.} Given a reference image $x^{+}$, we introduce moderate noise at level $t$:
\begin{equation}
x_t = (1{-}t)\, x^{+} + t\, \epsilon, \quad \epsilon \sim \mathcal{N}(0, I).
\label{eq:noise}
\end{equation}
The current generator then completes the remaining denoising from $x_t$ to obtain $x^{-}$, using one or two steps in practice. Reconstruction uses the conditioning associated with $x^{+}$: for text-to-image generation, this is a paired prompt or a caption; for image editing, it also includes the corresponding input images. With moderate noise ($t \le 0.5$), $x_t$ retains much of the reference structure, allowing the reconstruction to largely preserve style and scene layout. The remaining differences are thus more concentrated in fine details, encouraging the discriminator to assess fidelity rather than high-level content. In all three experimental settings, the noise level is randomly sampled from $\{0.3,0.5\}$ for each matched negative. Negatives are reconstructed online, so the pairs reflect the current generator's failure modes.

\textbf{Choosing reference images.} Fidelity references, or anchors, should cover the target visual quality and style, but need not maximize the task reward, which is optimized directly through $r_{\mathrm{spec}}$. We choose among three acquisition strategies according to visual compatibility and data availability:
\begin{itemize}
    \vspace{-1em}
    \item \emph{Base-model sampling.} When the base model's full-step outputs cover the target visual quality and style, we use them directly as anchors. This strategy is suitable when reward optimization primarily improves task correctness within a visual domain already covered by the base model. For GenEval, these are 50-step FLUX.1-dev outputs.
    \vspace{-0.2em}
    \item \emph{Target-data reuse.} Suitable existing target-domain images can be reused directly, regardless of base-model coverage. Sources include high-quality real images and existing high-quality generated images. When RL follows target-domain supervised fine-tuning (SFT), its ground-truth images can likewise be reused for both photorealistic and stylized tasks. For virtual dressing, real ground-truth try-on images from the training split serve as references.
    \vspace{-0.2em}
    \item \emph{Target-domain model sampling.} When base-model coverage is insufficient and suitable data are unavailable or disallowed by the experimental protocol, we sample a model that produces high-quality images in the target domain. An existing RL-aligned model is one possible source. Its samples can be prepared once and reused across runs with the same target visual requirements; the source model's scores and denoising trajectories are not used during AdvFidelity training. For HPDv2, we use 50-step samples from an HPSv2-aligned LeapAlign checkpoint.
\end{itemize}
All three strategies use the same content-matched pair construction. References may be stored beforehand or generated as needed; a fixed reference set is not required.

\subsection{Adversarial Training Procedure}
\label{sec:training}

\begin{algorithm}[t]
\caption{AdvFidelity Training}
\label{alg:training}
\begin{algorithmic}[1]
\REQUIRE pretrained policy $\pi_\theta$, discriminator $D_\phi$, task reward $r_{\mathrm{spec}}$, reference source $\mathcal{R}$, sampling budget $\tau$, noise level $t$, reward weight $\lambda$, discriminator update ratio $K$
\WHILE{not converged}
    \STATE Sample conditioning $c$ and rollouts $x \sim \pi_\theta(\cdot \mid c; \tau)$
    \STATE $r \leftarrow r_{\mathrm{spec}}(x, c) + \lambda \, D_\phi(x)$; update $\theta$ with the thresholded reward loss (Sec.~\ref{sec:overview})
    \FOR{$k = 1$ to $K$}
        \STATE Obtain a reference and its conditioning $(x^{+},c^{+})$ from $\mathcal{R}$
        \STATE Sample $\epsilon \sim \mathcal{N}(0,I)$; $x_t \leftarrow (1{-}t)\,x^{+} + t\,\epsilon$
        \STATE Complete denoising from $x_t$ with the current policy conditioned on $c^{+}$ to obtain $x^{-}$
        \STATE Update $\phi$ with the BT loss on $(x^{+},x^{-})$ (Eq.~\ref{eq:bt})
    \ENDFOR
\ENDWHILE
\end{algorithmic}
\end{algorithm}

Algorithm~\ref{alg:training} summarizes the alternating updates. In our FLUX experiments, the policy is initialized from the original FLUX.1-dev model. Each generator update optimizes both rewards on a few-step rollout from Gaussian noise. Each discriminator update obtains a reference from the chosen source and reconstructs a matched negative using the current generator. Preference alignment and few-step generation are therefore learned jointly, while reference acquisition can use existing images or model-generated samples.

Under ReFL, gradients of the discriminator score flow into the generator, inducing an adversarial interaction analogous to GAN training. Here, the discriminator supplies one reward within a multi-reward objective, and its training negatives are content-matched reconstructions rather than unrelated generations. Its role is to supervise the fidelity of few-step outputs as the task reward improves preference or task-specific performance.


\section{Experiments}
\label{exp}

We evaluate AdvFidelity on preference alignment with HPDv2 prompts, compositional generation with GenEval prompts, and a production-oriented virtual dressing task. The first two settings follow the experimental protocols of LeapAlign~\citep{leapalign}, including its training and evaluation prompt sets and evaluation procedures, and initialize AdvFidelity from the original FLUX.1-dev model. AdvFidelity jointly learns reward alignment and few-step generation in a single generator training stage, with the fidelity discriminator trained online from the chosen references. AdvFidelity uses 4 or 8 sampling steps for FLUX and 8 steps for virtual dressing, without classifier-free guidance (CFG). We compare against full-step LeapAlign RL trained on the same base model in each setting, including both CFG and non-CFG variants for virtual dressing. Together, these experiments evaluate the three reference acquisition strategies in Sec.~\ref{sec:pairs} across different visual domains and rewards.

\subsection{Preference Alignment on HPDv2}
\label{sec:exp-hpd}

\noindent \textbf{Implementation Details.}
We fine-tune FLUX.1-dev on the HPDv2 prompt set, which comprises 50,000 training prompts and 400 test prompts, using HPSv2~\citep{hpsv2} as the preference reward. We report HPSv2, PickScore~\citep{pickscore}, ImageReward~\citep{imagereward}, and the alignment and image-quality sub-scores of UnifiedReward~\citep{unifiedreward}. Following LeapAlign, the table labels HPSv2 as an in-domain metric and PickScore, ImageReward, and UnifiedReward as out-of-domain metrics. HPSv3~\citep{hpsv3} is shaded and reported separately for reference only, as it can assign misleading scores to images with few-step artifacts; Appendix~\ref{sec:supp-metric-reliability} examines artifact sensitivity on content-matched pairs.

We compare against DanceGRPO~\citep{dancegrpo}, MixGRPO~\citep{mixgrpo}, ReFL~\citep{imagereward}, DRTune~\citep{drtune}, DRaFT-LV~\citep{draft}, and LeapAlign~\citep{leapalign}. For HPDv2 and GenEval, results for these methods and the pretrained baseline are taken from LeapAlign. RTDMD$^{*}$ denotes the officially released four-step FLUX.1-dev RTDMD~\citep{rtdmd} checkpoint evaluated at 4 and 8 NFEs. The public FLUX.1-dev RTDMD configurations use 25,432 PickScore prompts for DMD warm-up and 50,000 GenEval prompts with CLIPScore, HPSv2, PickScore, and GenEval rewards for RL. For a fairer comparison, we also retrain RTDMD using its released code at 4 and 8 steps with the same task training prompts and HPSv2 reward as AdvFidelity under LeapAlign's protocol; unmarked RTDMD entries report these runs.

The fidelity discriminator consists of a frozen DINOv2-Large~\citep{dinov2} encoder and a trainable MLP head. We use crop sizes of 256 and 384 pixels with a crop-level loss weight of 0.3. The generator and discriminator are optimized with AdamW at learning rates of $1\times10^{-4}$ and $5\times10^{-5}$, respectively. The gradient discount factor is 0.3, the reward margin $m$ is 0.55, and the fidelity reward weight $\lambda$ is 0.05. The discriminator performs four updates per generator update.

\begin{table}[t]
\caption{Quantitative comparison of preference alignment on FLUX.1-dev. \textbf{Bold} and \underline{underlined} values denote the best and second best results. Higher is better. $^\dagger$HPSv3 is shaded and reported for reference only, as it assigns unreliable scores to artifact-laden images. $^{*}$Officially released RTDMD models; unmarked RTDMD rows are retrained under the same protocol as AdvFidelity.}
\label{tab:flux-hpd-hps}
\begin{center}
\begin{tabular}{lccccccc}
\toprule
& &\multicolumn{1}{c}{\bf In-Domain} &\multicolumn{4}{c}{\bf Out-of-Domain} & \\
\cmidrule(lr){3-3} \cmidrule(lr){4-7}
\multicolumn{1}{l}{\bf Methods} &\multicolumn{1}{c}{\bf NFE} &\multicolumn{1}{c}{\bf HPSv2} &\multicolumn{1}{c}{\bf PickScore} &\multicolumn{1}{c}{\bf ImgRwd} &\multicolumn{1}{c}{\bf UR-Align} &\multicolumn{1}{c}{\bf UR-IQ} &\multicolumn{1}{c}{\bf HPSv3$^\dagger$} 
\\ 
\midrule
\multicolumn{8}{c}{\textit{Pre-trained Models}} \\
\midrule
FLUX.1-dev           &50 &0.3083 &22.6970 &1.0210 &3.4428 &3.4792 &\cellcolor{gray!15}13.3406 \\
\midrule
\multicolumn{8}{c}{\textit{RL Methods}} \\
\midrule
DanceGRPO            &50 &0.3451 &23.1186 &1.2347 &3.4660 &3.6199 &\cellcolor{gray!15}14.8336 \\
MixGRPO              &50 &0.3692 &23.5184 &\textbf{1.6155} &3.4393 &3.6241 &\cellcolor{gray!15}14.7530 \\
ReFL                 &50 &0.3852 &23.6299 &1.3468 &3.4786 &3.6870 &\cellcolor{gray!15}15.5127 \\
DRTune               &50 &0.3882 &23.5185 &1.3384 &3.4793 &3.6679 &\cellcolor{gray!15}15.5606 \\
DRaFT-LV             &50 &0.3859 &23.6437 &1.3562 &3.4868 &3.6887 &\cellcolor{gray!15}15.3699 \\
LeapAlign            &50 &\underline{0.4096} &\underline{23.7239} &1.5105 &3.4837 &3.7284 &\cellcolor{gray!15}15.7548 \\
\midrule
\multicolumn{8}{c}{\textit{Joint Methods}} \\
\midrule
RTDMD*                &4   &0.3352  &22.7344  &1.2304  &3.4243  &3.6373  &\cellcolor{gray!15}14.6269   \\
RTDMD*                &8   &0.3283  &22.4196  &1.0885  &3.4167  &3.6769  &\cellcolor{gray!15}14.5304  \\
RTDMD               &4  &0.3724 &22.8279 &1.3020 &3.4015 &3.7452 &\cellcolor{gray!15}15.1737 \\
RTDMD                &8  &0.3923 &23.3059 &1.4134 &3.4779 &\underline{3.7482} &\cellcolor{gray!15}15.0931 \\
AdvFidelity              &4  &0.4055 &23.6211 &1.5164 &\underline{3.4981} &3.7338 &\cellcolor{gray!15}15.4299 \\
AdvFidelity              &8  &\textbf{0.4125} &\textbf{23.7685} &\underline{1.5255} &\textbf{3.5024} &\textbf{3.7509} &\cellcolor{gray!15}15.3197 \\
\bottomrule
\end{tabular}
\end{center}
\end{table}

\begin{table}[h]
\caption{Compositional generation results on FLUX.1-dev, with HPSv2 as the training reward. \textbf{Bold} and \underline{underlined} values denote the best and second best results. Higher is better. $^{*}$Officially released RTDMD models; unmarked RTDMD rows are retrained under the same protocol as AdvFidelity.}
\label{tab:flux-geneval-hps}
\begin{center}
\begin{tabular}{lcccccccc}
\toprule
\multicolumn{1}{l}{\bf Methods} &\multicolumn{1}{c}{\bf NFE} &\multicolumn{1}{c}{\bf Overall} &\multicolumn{1}{c}{\bf Single Obj} &\multicolumn{1}{c}{\bf Two Obj} &\multicolumn{1}{c}{\bf Count} &\multicolumn{1}{c}{\bf Color} &\multicolumn{1}{c}{\bf Pos} &\multicolumn{1}{c}{\bf AttrB}
\\ 
\midrule
\multicolumn{9}{c}{\textit{Pre-trained Models}} \\
\midrule
FLUX.1-dev           &50 &0.6481 &99.38 &85.10 &66.56 &76.06 &19.00 &42.75 \\
\midrule
\multicolumn{9}{c}{\textit{RL Methods}} \\
\midrule
DanceGRPO            &50 &0.6775 &99.38 &90.15 &69.38 &76.33 &22.25 &49.00 \\
MixGRPO              &50 &0.7232 &\underline{99.69} &93.69 &\underline{80.00} &80.05 &24.25 &56.25 \\
ReFL                 &50 &0.7011 &99.38 &92.68 &69.06 &75.80 &26.75 &57.00 \\
DRTune               &50 &0.7101 &99.38 &93.69 &73.12 &76.86 &27.50 &55.50 \\
DRaFT-LV             &50 &0.7024 &\underline{99.69} &92.42 &74.06 &75.53 &24.00 &55.75 \\
LeapAlign            &50 &0.7420 &99.38 &\underline{96.46} &72.50 &80.59 &30.25 &66.00 \\
\midrule
\multicolumn{9}{c}{\textit{Joint Methods}} \\
\midrule
RTDMD*                &4  &0.6521  &97.50  &85.86  &53.44  &78.19  &26.75  &49.50  \\
RTDMD*                &8  &0.6828  &\underline{99.69}  &91.41  &63.44  &73.14  &25.50  &56.50  \\
RTDMD                &4  &0.6794 &99.38 &92.68 &73.75 &63.83 &\underline{33.25} &44.75 \\
RTDMD                &8  &0.7236 &\underline{99.69} &95.45 &72.81 &74.73 &\textbf{36.25} &55.25 \\
AdvFidelity              &4  &\underline{0.7535} &99.38 &\underline{96.46} &74.06 &\underline{82.45} &30.75 &\underline{69.00} \\
AdvFidelity              &8  &\textbf{0.7759} &\textbf{100.0} &\textbf{98.99} &\textbf{82.50} &\textbf{84.57} &26.75 &\textbf{72.75} \\
\bottomrule
\vspace{-3em}
\end{tabular}
\end{center}
\end{table}

\noindent \textbf{Reference Images and Matched Negatives.}
We use \emph{target-domain model sampling} (Sec.~\ref{sec:pairs}). In this setting, HPSv2 optimization encourages pronounced cartoon and fantasy styles, substantially shifting the visual distribution from that of the original FLUX.1-dev model. Fidelity references should cover this preferred appearance. Although HPDv2 provides approximately 430,000 images from multiple generative models, this heterogeneous collection does not consistently reflect the reward-preferred appearance. We therefore prepare 50,000 reference image--prompt pairs once, using 50-step samples from the official HPSv2-aligned LeapAlign~\citep{leapalign} checkpoint, and use them directly without filtering by any reward score. Following LeapAlign's HPSv2 scoring procedure, we obtain mean scores of 0.2583 for the released HPDv2 images and 0.3881 for these generated references, providing a quantitative comparison of their compatibility with the training reward. This reference set can be reused across training runs with the same target visual requirements, while the fidelity discriminator adapts online to each generator. AdvFidelity itself starts from the original FLUX.1-dev model and does not use the reference model's scores or denoising trajectories.

Across HPDv2, GenEval, and virtual dressing, the noise level is randomly sampled from $\{0.3,0.5\}$ when constructing each matched negative. The current generator completes the remaining denoising in one or two steps.

\begin{figure}[t]
\begin{center}
\includegraphics[width=1\linewidth]{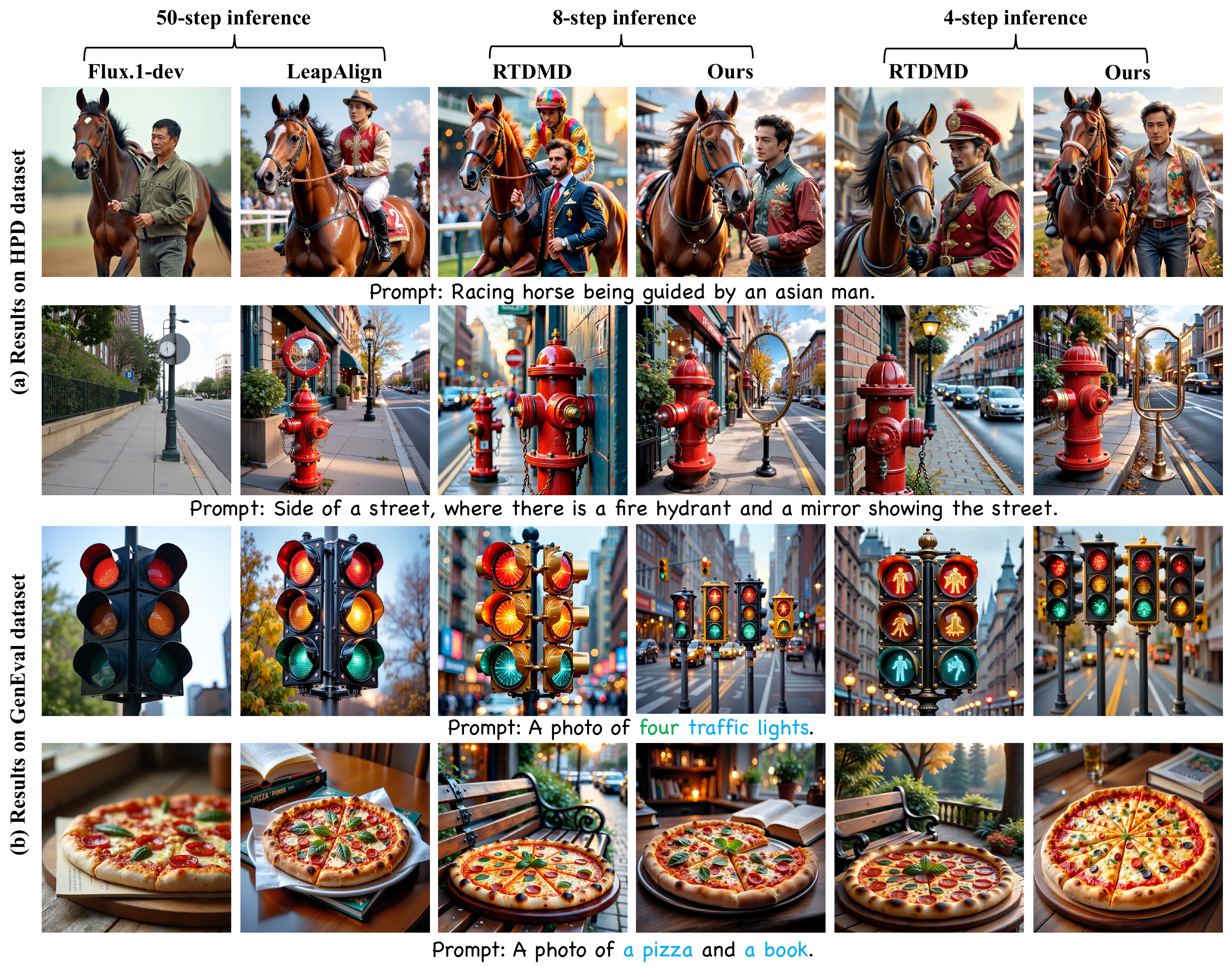}
\vspace{-2em}
\end{center}
\caption{Qualitative comparisons on (a) HPDv2 preference-alignment prompts and (b) GenEval compositional prompts. Full-step FLUX.1-dev and LeapAlign are compared with 4- and 8-step RTDMD and AdvFidelity. Please zoom in for artifact details}
\label{fig:qual}
\end{figure}

\noindent \textbf{Results.}
Table~\ref{tab:flux-hpd-hps} shows that 8-step AdvFidelity achieves higher scores than 50-step LeapAlign on the five primary metrics, with HPSv2 increasing from 0.4096 to 0.4125. Among the evaluated 50-step RL methods, 8-step AdvFidelity leads on HPSv2, PickScore, UR-Align, and UR-IQ, while MixGRPO retains the highest ImageReward. At 4 steps, its HPSv2 and PickScore remain close to those of full-step LeapAlign, while ImageReward, UR-Align, and UR-IQ are higher. Compared with retrained RTDMD at the same sampling budgets, AdvFidelity achieves higher HPSv2, PickScore, ImageReward, and UR-Align scores at both 4 and 8 steps. It also exceeds the released RTDMD$^{*}$ checkpoint on all five primary metrics at both sampling budgets. Figure~\ref{fig:qual}(a) shows substantially reduced few-step artifacts relative to RTDMD, while retaining the preferred appearance and fine details of full-step RL outputs. Together, the quantitative and qualitative results support overall visual quality comparable to full-step RL with far fewer sampling steps.

\subsection{Compositional Generation on GenEval}
\label{sec:exp-geneval}

\noindent \textbf{Implementation Details and Reference Images.}
We evaluate compositional generation on GenEval~\citep{geneval}, using HPSv2~\citep{hpsv2} as the preference reward. GenEval assesses object presence, counting, color, spatial relations, and attribute binding in familiar visual scenes. Training uses 50,000 prompts, and evaluation follows the standard protocol with 553 test prompts, generating four images per prompt. Although both settings optimize HPSv2, we observe much milder changes in visual style on GenEval prompts than on HPDv2 prompts. The original FLUX.1-dev model's 50-step outputs therefore remain suitable fidelity references, allowing us to use \emph{base-model sampling} (Sec.~\ref{sec:pairs}) without an RL-aligned reference model. These references supervise fidelity and need not satisfy every compositional constraint; the generator is optimized directly with HPSv2. Therefore, we prepare 5,000 reference image--prompt pairs using 50-step FLUX.1-dev inference.
Matched-negative construction and the remaining training settings follow Sec.~\ref{sec:exp-hpd}. We report the overall GenEval score and its sub-category scores.

\noindent \textbf{Results.}
Table~\ref{tab:flux-geneval-hps} shows overall scores of 0.7535 and 0.7759 for 4- and 8-step AdvFidelity, respectively, compared with 0.7420 for 50-step LeapAlign. Both variants exceed all evaluated 50-step RL baselines on the overall score and improve color and attribute binding over LeapAlign. They also outperform both retrained RTDMD and released RTDMD$^{*}$ at the corresponding sampling budgets on the overall score. These results show that base-model references can provide effective fidelity supervision while the generator learns improved compositional performance and few-step generation jointly. Figure~\ref{fig:qual}(b) provides qualitative examples at both sampling budgets.

\subsection{Virtual Dressing on Real Data}
\label{sec:exp-dressing}

\begin{table}[t]
\caption{Quantitative comparison on virtual dressing with Qwen-Image-Edit-2511.}
\label{tab:qwen-tryon}
\begin{center}
\setlength{\tabcolsep}{2pt}
\begin{tabular}{lccccccc}
\toprule
& & &\multicolumn{1}{c}{\bf In-Domain} &\multicolumn{4}{c}{\bf Out-of-Domain} \\
\cmidrule(lr){4-4} \cmidrule(lr){5-8}
\multicolumn{1}{l}{\bf Methods} &\multicolumn{1}{c}{\bf CFG} &\multicolumn{1}{c}{\bf NFE} &\multicolumn{1}{c}{\bf DINO} &\multicolumn{1}{c}{\bf Realism} &\multicolumn{1}{c}{\bf Consistency} &\multicolumn{1}{c}{\bf Alignment} &\multicolumn{1}{c}{\bf Coherence} \\
\midrule
Qwen-Image-Edit-2511    &$\checkmark$ &40 &0.7426 &5.96${\scriptstyle \pm 0.86}$ &8.60${\scriptstyle \pm 0.49}$ &8.77${\scriptstyle \pm 0.45}$ &7.91${\scriptstyle \pm 0.51}$ \\
\midrule
LeapAlign             &$\times$ &40 &0.7793 &8.06${\scriptstyle \pm 0.62}$ &9.35${\scriptstyle \pm 0.40}$ &9.03${\scriptstyle \pm 0.44}$ &8.19${\scriptstyle \pm 0.49}$ \\
LeapAlign             &$\checkmark$ &40 &0.7874 &8.33${\scriptstyle \pm 0.42}$ &9.28${\scriptstyle \pm 0.42}$ &9.06${\scriptstyle \pm 0.39}$ &8.30${\scriptstyle \pm 0.46}$ \\
\midrule
AdvFidelity               &$\times$ &8  &\textbf{0.7929} &\textbf{8.45}${\scriptstyle \pm 0.36}$ &\textbf{9.54}${\scriptstyle \pm 0.27}$ &\textbf{9.11}${\scriptstyle \pm 0.45}$ &\textbf{8.32}${\scriptstyle \pm 0.42}$ \\
\bottomrule
\end{tabular}
\end{center}
\end{table}

\begin{figure}[t]
\begin{center}
\includegraphics[width=1\linewidth]{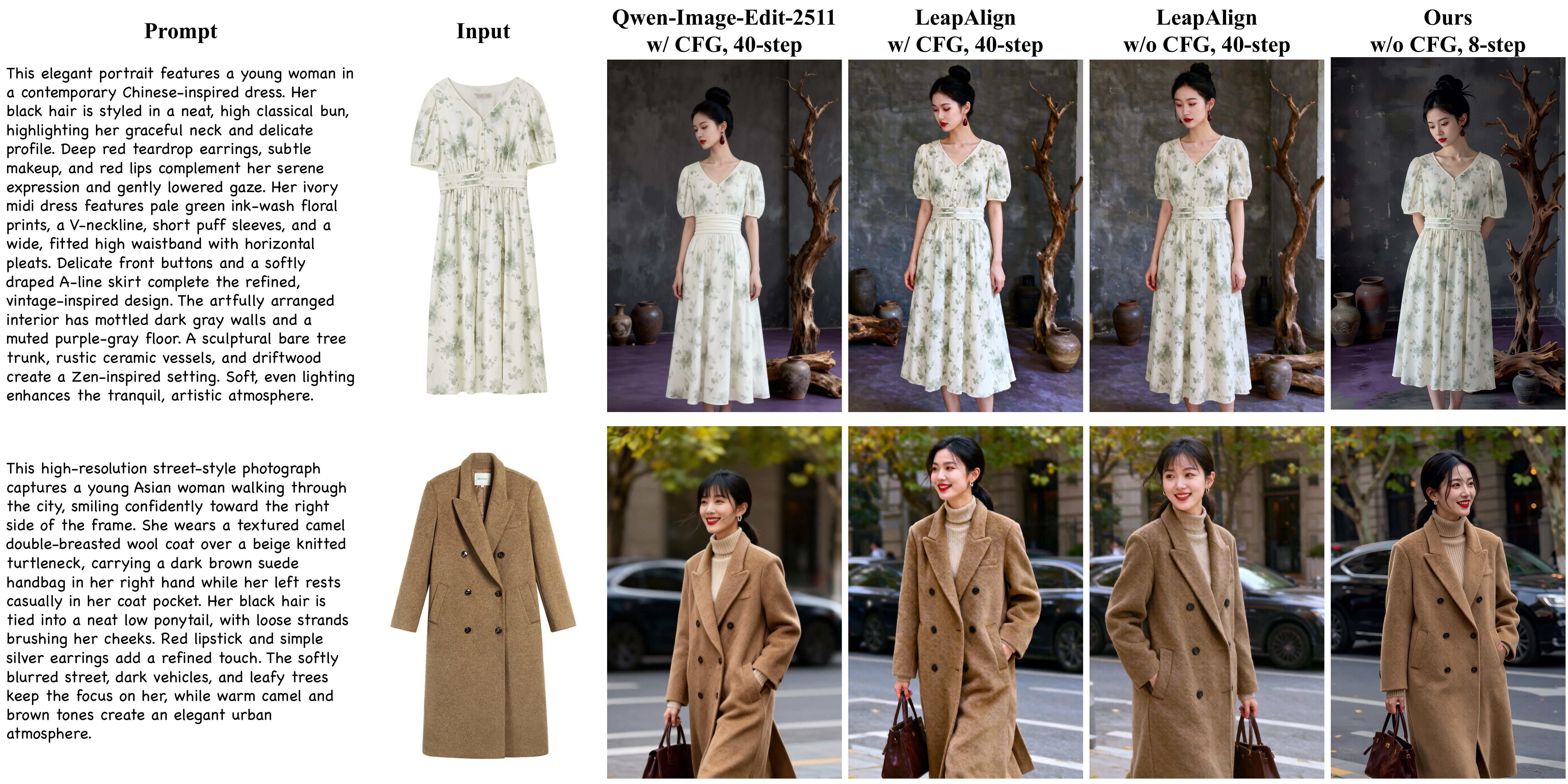}
\end{center}
\caption{Qualitative comparison on virtual dressing with Qwen-Image-Edit-2511. AdvFidelity uses 8 steps without CFG; the full-step comparisons use 40 steps, with CFG settings indicated above each column. Please zoom in for artifact details}
\vspace{-2mm}
\label{fig:qwen}
\end{figure}

\noindent \textbf{Implementation Details and Reference Images.}
We evaluate AdvFidelity on virtual dressing using Qwen-Image-Edit-2511~\citep{qwen-image}. The task takes a prompt and a reference garment image as input, and generates an image that presents the garment as described by the prompt. The task data consist of outfit images collected from the web, comprising 1,000 training pairs and 124 test samples. We use \emph{target-data reuse} (Sec.~\ref{sec:pairs}): real ground-truth try-on images drawn exclusively from the training split of the paired task data serve directly as fidelity references. Suitable real images are often available in industrial applications; when RL follows target-domain SFT, the SFT ground-truth images can likewise be reused as references. Our dressing setting illustrates this use of existing target data without an RL-aligned reference model or additional reference-image generation. Matched negatives are reconstructed online by the current generator, and the discriminator architecture and objective follow Sec.~\ref{sec:exp-hpd}. The generator and discriminator are both optimized with AdamW at a learning rate of $5\times10^{-5}$, the reward margin $m$ is 1, the fidelity reward weight $\lambda$ is 0.01, and the discriminator performs two updates per generator update. To encourage garment consistency and realism, we use DINO feature similarity between each generated image and its ground truth as the task-specific reward. Thus, training ground-truth images serve both as fidelity references and as comparison targets for this reward. We report DINO similarity and use Gemini 3.5 Flash to evaluate Realism, Consistency, Alignment, and Coherence on the 124 test samples, each on an integer scale from 1 to 10. We repeat Gemini evaluation three times on the same generated images to reduce the effect of stochastic MLLM scoring, and report the mean $\pm$ standard deviation across the three runs.

\noindent \textbf{Results.}
Table~\ref{tab:qwen-tryon} shows that 8-step AdvFidelity without CFG achieves the highest DINO similarity and the highest mean Gemini scores across all four evaluation aspects among the compared methods. Its DINO similarity is 0.7929, compared with 0.7874 for 40-step LeapAlign with CFG. These results support the applicability of AdvFidelity to a practical image-editing task with real-image references and a DINO similarity reward for garment consistency and realism, enabling few-step inference without CFG. Figure~\ref{fig:qwen} provides qualitative comparisons; the displayed prompts are shortened for readability, with full prompts shown in Figure~\ref{fig:supp-tryon}.

\subsection{Ablation Studies}
\label{sec:exp-ablation}

We evaluate four cumulative configurations of AdvFidelity at 8-step inference on HPDv2, together with two comparison baselines. The original FLUX.1-dev model sampled in 8 steps is the pretrained baseline. LeapAlign+DMD distills the official full-step LeapAlign checkpoint to 8 steps and represents sequential RL followed by distillation. The vanilla AdvFidelity configuration uses a frozen discriminator trained on references from \emph{base-model sampling} and unpaired rollout negatives. We then add online discriminator updates (+adaptive $r_{\mathrm{fid}}$), replace the references with images obtained through \emph{target-domain model sampling} (+target-anchored positives), and introduce content-matched negatives to obtain the full method. The Base and Target labels in Table~\ref{tab:ablation} denote these two reference acquisition strategies, respectively.

\begin{table}[t]
\caption{Ablation study at 8-step inference under HPDv2 setting. \textbf{Bold} values denote the best results.}
\vspace{-2mm}
\label{tab:ablation}
\begin{center}
\resizebox{\linewidth}{!}{%
\begin{tabular}{@{}lcccccccc@{}}
\toprule
\multicolumn{1}{l}{\bf Methods} &\multicolumn{1}{c}{\bf Paired} &\multicolumn{1}{c}{\bf Anchor} &\multicolumn{1}{c}{\bf $r_{fid}$} &\multicolumn{1}{c}{\bf HPSv2} &\multicolumn{1}{c}{\bf PickScore} &\multicolumn{1}{c}{\bf ImgRwd} &\multicolumn{1}{c}{\bf UR-Align} &\multicolumn{1}{c}{\bf UR-IQ} \\
\midrule
FLUX.1-dev &-- &-- &-- &0.2929 &22.3912 &0.9532 &3.4227 &3.7024  \\
LeapAlign + DMD &-- &-- &-- &0.3978 &22.9654 &1.3976 &3.4259 &3.7164  \\
\midrule
Vanilla &$\times$ &Base &Frozen &0.3416 &22.6735 &1.1650 &3.4404 &3.7123  \\
+Adaptive $r_{\mathrm{fid}}$ &$\times$ &Base &$\checkmark$ &0.3558 &22.8637 &1.2171 &3.4494 &3.7215  \\
+Target-anchored positives &$\times$ &Target &$\checkmark$ &0.3920 &23.2588 &1.3567 &3.4456 &3.7141  \\
\midrule
+Paired Data (Full)  &$\checkmark$ &Target &$\checkmark$ &\textbf{0.4125} &\textbf{23.7685} &\textbf{1.5255} &\textbf{3.5024} &\textbf{3.7509}  \\
\bottomrule
\vspace{-2em}
\end{tabular}}
\end{center}
\end{table}

\begin{figure}[t]
\begin{center}
\includegraphics[width=1\linewidth]{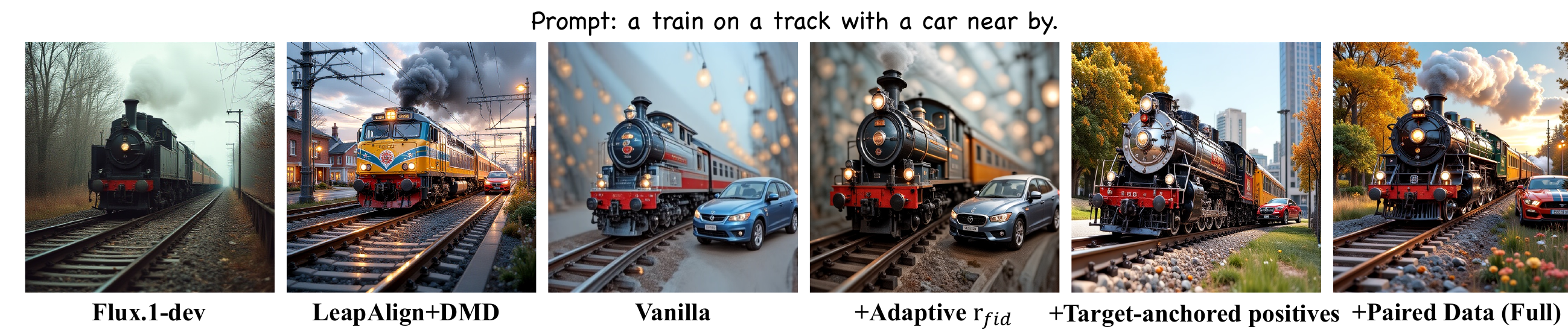}
\end{center}
\caption{Qualitative ablation comparison on FLUX.1-dev with HPDv2 prompts at 8-step inference. Components are added cumulatively from the vanilla configuration to the full method.}
\vspace{-1mm}
\label{fig:ablation}
\end{figure}

\noindent \textbf{Results.}
Sequential DMD lowers the full-step LeapAlign checkpoint's HPSv2 score from 0.4096 to 0.3978 and PickScore from 23.7239 to 22.9654, illustrating a loss of alignment gains in this configuration. Online discriminator updates improve all five reported metrics over the frozen-discriminator variant, supporting the benefit of adapting fidelity supervision to the current generator. Switching to target-style references further raises HPSv2 from 0.3558 to 0.3920, consistent with improved reference compatibility in this stylized setting. Adding content-matched negatives then improves all five metrics, with the full method achieving the strongest results among the ablation configurations. Figure~\ref{fig:ablation} illustrates progressively reduced grid-like artifacts as the components are added. Together, these results support online fidelity supervision, suitable reference selection, and content-matched reconstruction within the joint training framework.


\section{Conclusion}

We presented AdvFidelity, a framework that jointly learns reward alignment and few-step generation within a single RL stage, without distillation warm-up or DMD. It builds on the observation that reward optimization on few-step trajectories improves structural coherence, while low-level artifacts may escape task-reward supervision. AdvFidelity therefore complements the task reward with an online fidelity discriminator trained on content-matched pairs that reduce high-level differences and encourage attention to low-level artifacts. Through LeapAlign, both rewards are optimized along the few-step sampling trajectories used at inference. Experiments across two model backbones and three task settings show that, using only 4 or 8 steps without CFG, AdvFidelity matches full-step LeapAlign in alignment, task performance, and overall generation quality and remains highly competitive with other evaluated full-step RL methods.

\clearpage


\bibliography{iclr2027_conference}
\bibliographystyle{iclr2027_conference}

\clearpage
\appendix

\section{Additional Training Details}
\label{sec:supp-training}

\textbf{Training configuration.} Table~\ref{tab:supp-training} summarizes the training iterations, effective global batch size, image resolution, and generator LoRA rank of each experimental setting. Every variant is trained at the sampling budget used at inference; the 4-step and 8-step FLUX variants are separately trained models, each using the settings shown for its experimental setting.

\textbf{Generator LoRA.} The public LeapAlign FLUX implementation fine-tunes all DiT parameters. We use low-rank adaptation (LoRA) due to GPU memory constraints. In all three settings, adapters are applied to both attention and MLP projections while the remaining base-model weights are kept frozen; their ranks are listed in Table~\ref{tab:supp-training}. The discriminator backbone is also frozen, and only its MLP head is trained (Sec.~\ref{sec:reward}).

\begin{table}[h]
\caption{Training configuration of each experimental setting. Batch size is the effective global value (per-device batch size $\times$ gradient accumulation $\times$ 8 GPUs); iterations are counted after gradient accumulation.}
\label{tab:supp-training}
\begin{center}
\begin{tabular}{lcccc}
\toprule
Setting &Iterations &Effective batch size &Resolution &LoRA rank \\
\midrule
HPDv2 &450 &64 &$720\times720$ &256 \\
GenEval &500 &64 &$720\times720$ &256 \\
Virtual dressing &750 &16 &$1024\times1024$ &128 \\
\bottomrule
\end{tabular}
\end{center}
\end{table}

\section{Reliability of Automated Metrics on Few-Step Artifacts}
\label{sec:supp-metric-reliability}

This appendix examines whether commonly used automated metrics can identify artifacts caused by short denoising trajectories. The analysis is conducted in the FLUX.1-dev HPDv2 setting (Sec.~\ref{sec:exp-hpd}), following the content-matched pair-construction procedure used for fidelity-reward training (Sec.~\ref{sec:pairs}). Each positive is a 50-step sample from the HPSv2-aligned LeapAlign checkpoint that serves as a fidelity reference. To obtain its negative, we add moderate forward noise and reconstruct the reference in one or two denoising steps with the original FLUX.1-dev model. This base model has not been optimized for few-step inference, so the short reconstruction introduces visible artifacts. Moderate noising retains much of the reference content and structure, concentrating pair differences in low-level artifacts. We collect 1,000 pairs (2,000 images) and score every image with six widely used metrics: HPSv3, HPSv2, Aesthetic Score, PickScore, ImageReward, and CLIPScore. Text-conditioned metrics are computed with the prompt of the corresponding pair.

\begin{figure}[h]
\begin{center}
\includegraphics[width=1\linewidth]{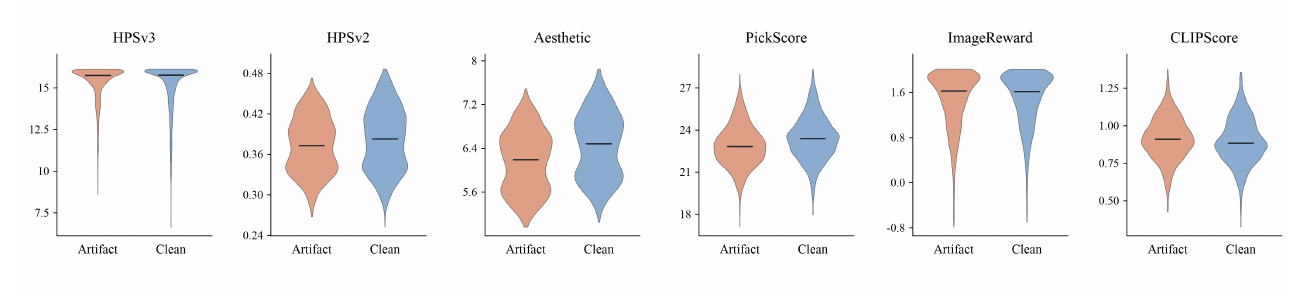}
\end{center}
\caption{Score distributions of six automated metrics on reference images and their short reconstructions by the original FLUX.1-dev model, from 1,000 content-matched pairs.}
\label{fig:violin}
\end{figure}

Figure~\ref{fig:violin} compares the score distributions on reference positives and artifact-laden negatives. The width of each violin shows the density of scores, and the black bar marks the median. HPSv3 and HPSv2 assign higher median scores to artifact-laden images than to the reference positives. The other metrics show substantial overlap between the two groups. Because marginal distributions do not show how each matched pair is ranked, we examine paired differences below.

\textbf{Pairwise score differences and positive win rates.} To test whether each metric ranks a reference above its matched reconstruction, we compare scores within the same 1,000 pairs. Let $s_i^+$ and $s_i^-$ be the positive and negative scores for pair $i$, respectively. We report the mean paired difference $\overline{\Delta}=N^{-1}\sum_{i=1}^{N}(s_i^+-s_i^-)$ and the positive win rate, the percentage of pairs with $s_i^+>s_i^-$, where $N=1{,}000$. A positive mean favors references on average, while a win rate above $50\%$ means they score higher in most pairs. There are no tied scores in these pairs.

\begin{table}[h]
\caption{Pairwise comparison on 1,000 content-matched positive--negative pairs. Mean $\Delta$ is the reference score minus the reconstruction score, averaged over pairs; positive win rate is the percentage of pairs in which the reference scores higher.}
\label{tab:supp-metric-pairs}
\begin{center}
\begin{tabular}{lrr}
\toprule
Metric & Mean $\Delta$ & Positive win rate (\%) \\
\midrule
HPSv3 & $-0.1748$ & 38.30 \\
HPSv2 & $-0.0087$ & 79.90 \\
Aesthetic Score & $+0.2955$ & 95.60 \\
PickScore & $+0.5341$ & 83.30 \\
ImageReward & $+0.0070$ & 48.20 \\
CLIPScore & $-0.0152$ & 40.30 \\
\bottomrule
\end{tabular}
\end{center}
\end{table}

Table~\ref{tab:supp-metric-pairs} shows that HPSv3 ranks the reference higher in only $38.30\%$ of pairs, with a mean paired difference of $-0.1748$; it favors the artifact-laden reconstruction in the remaining $61.70\%$. HPSv2 favors the reference in $79.90\%$ of pairs. Despite this positive win rate, its mean paired difference is $-0.0087$, indicating that the reverse-score margins in the remaining pairs outweigh the positive margins. Aesthetic Score and PickScore favor references more often ($95.60\%$ and $83.30\%$), while ImageReward and CLIPScore yield positive win rates of $48.20\%$ and $40.30\%$. The metrics therefore vary substantially in their sensitivity to these artifacts and do not uniformly provide reliable fidelity supervision, consistent with Figure~\ref{fig:motivation-b}. HPSv2 remains the in-domain training reward and is reported as such in Table~\ref{tab:flux-hpd-hps}, but its score alone is not used as evidence of artifact removal. HPSv3 is shaded and reported for reference only because it frequently favors the artifact-laden reconstruction in these matched pairs.

\section{Additional Qualitative Results}
\label{sec:supp-vis}

\begin{figure}[t]
\begin{center}
\includegraphics[width=1\linewidth]{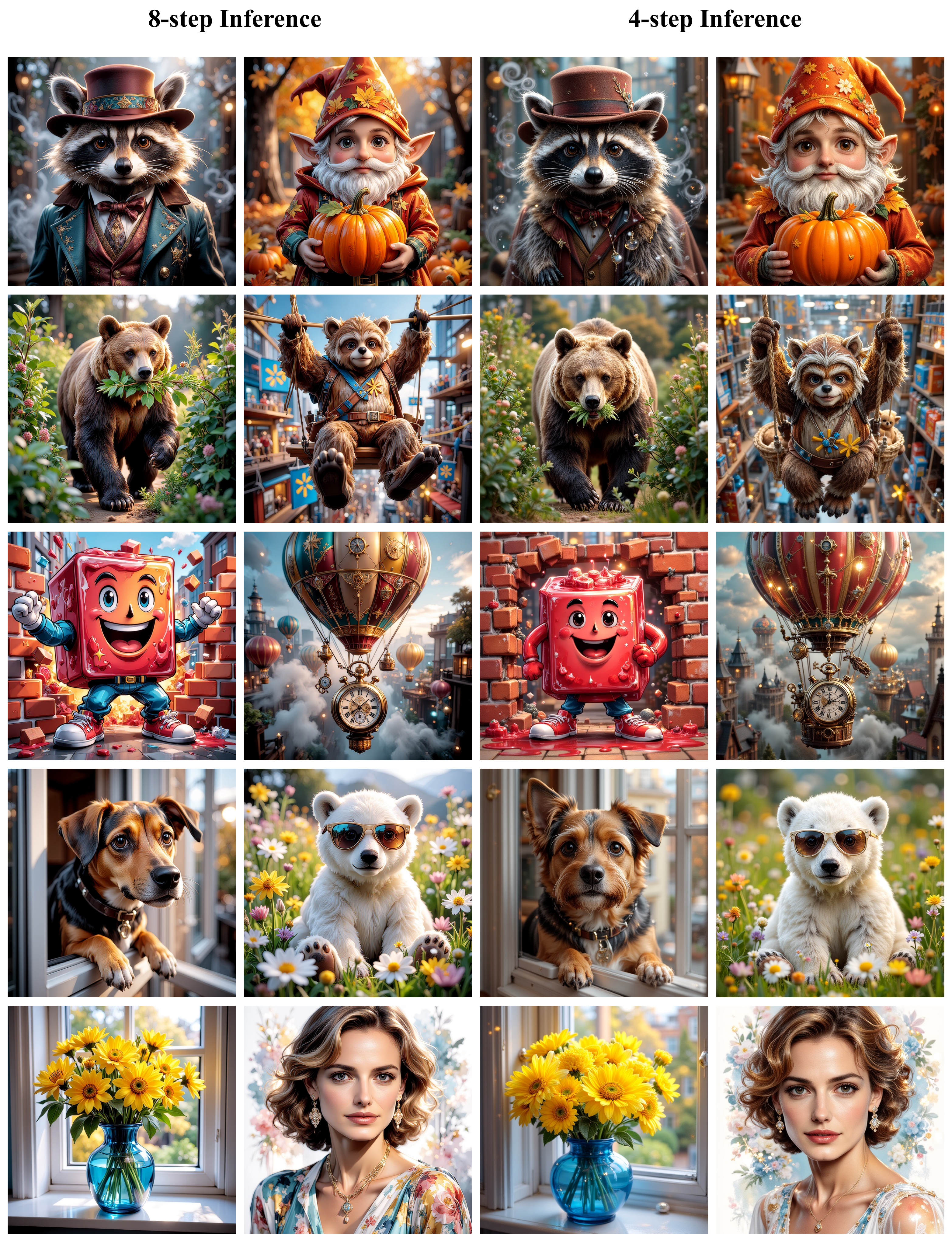}
\end{center}
\caption{Additional qualitative results on HPDv2 prompts (left: 8-step, right: 4-step).}
\label{fig:supp-hpd}
\end{figure}

\begin{figure}[t]
\begin{center}
\includegraphics[width=1\linewidth]{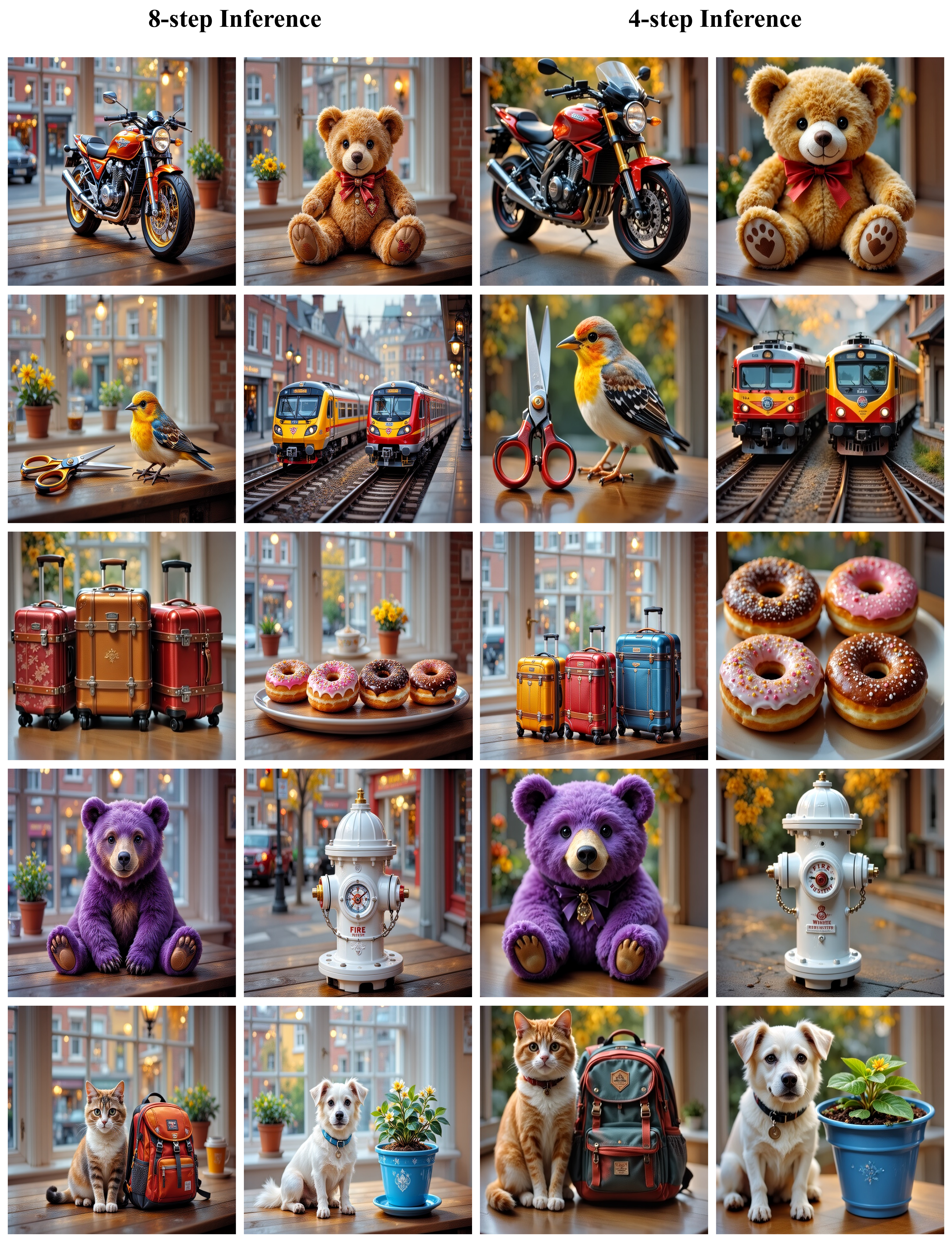}
\end{center}
\caption{Additional qualitative results on GenEval prompts (left: 8-step, right: 4-step).}
\label{fig:supp-geneval}
\end{figure}

\begin{figure}[t]
\begin{center}
\includegraphics[width=1\linewidth]{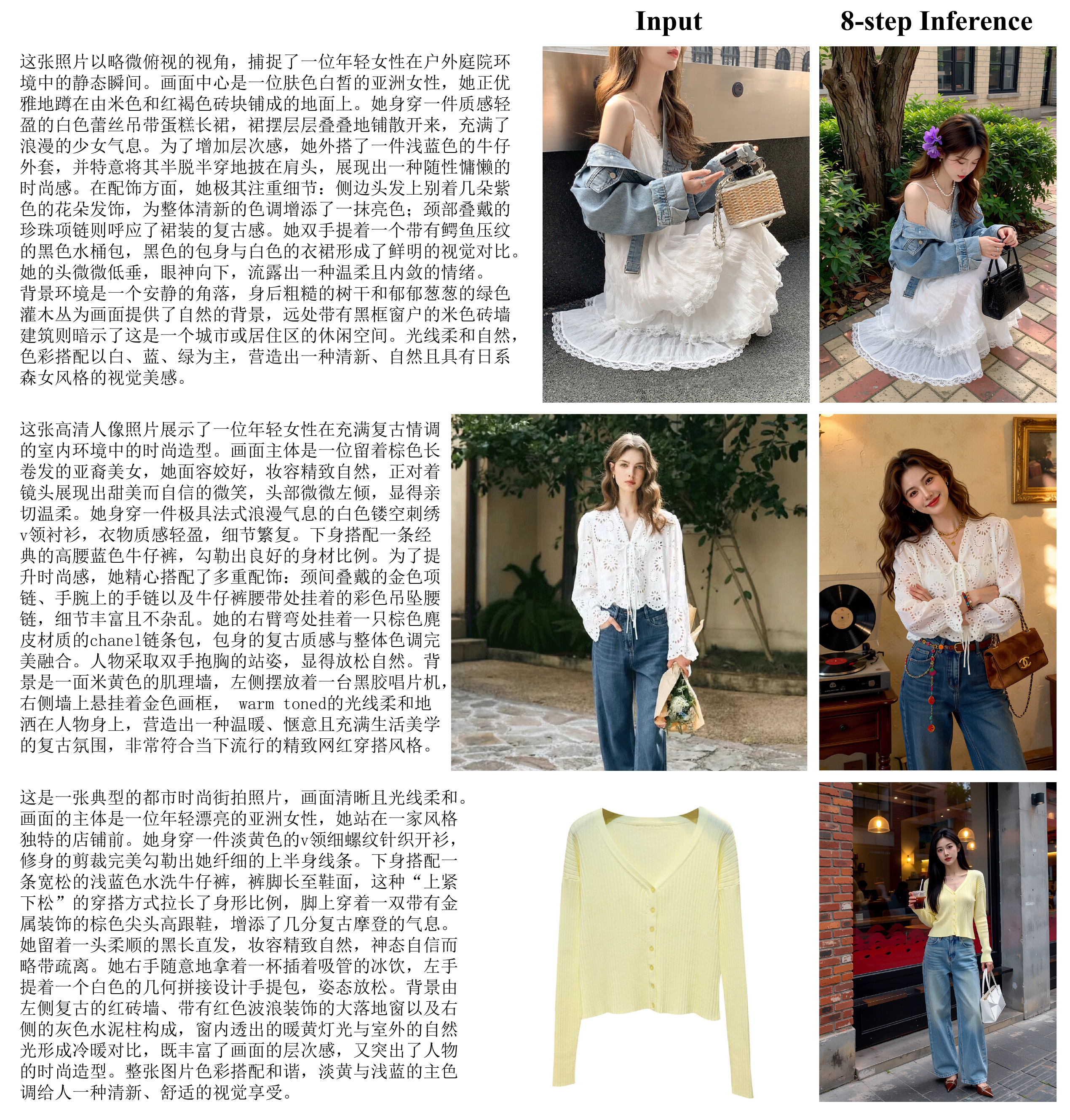}
\end{center}
\caption{Additional qualitative results on virtual dressing (left: prompt, mid: input garment, right: 8-step).}
\label{fig:supp-tryon}
\end{figure}

We provide additional qualitative results of AdvFidelity in the three experimental settings: HPDv2 preference alignment (Figure~\ref{fig:supp-hpd}), GenEval compositional generation (Figure~\ref{fig:supp-geneval}), and virtual dressing (Figure~\ref{fig:supp-tryon}). The HPDv2 and GenEval figures show 8-step generations on the left and 4-step generations on the right. The virtual dressing figure shows prompts, input garments, and 8-step generations.

\end{document}